%% file: neurips_2023.tex
\documentclass{article}

\PassOptionsToPackage{numbers, compress}{natbib}

\usepackage[preprint]{neurips_2023}

\usepackage[utf8]{inputenc} 
\usepackage[T1]{fontenc}    
\usepackage{hyperref}       
\usepackage{url}            
\usepackage{booktabs}       
\usepackage{amsfonts}       
\usepackage{nicefrac}       
\usepackage{microtype}      
\usepackage{xcolor}         

\usepackage{times}
\usepackage{latexsym}
\usepackage{enumitem}

\usepackage{graphicx}
\usepackage{amssymb}
\usepackage{comment}

\usepackage{caption}
\usepackage{subcaption}
\usepackage{booktabs}

\usepackage{multicol}
\usepackage{multirow}
\usepackage{array, tabularx,  ragged2e,  booktabs}
\usepackage{natbib}

\usepackage[utf8]{inputenc}
\usepackage{graphicx}
\usepackage{pgfplots}
\pgfplotsset{compat=1.14}

\usepackage{ctable}
\usepackage{wrapfig}
\usepackage{tikz}
\usetikzlibrary{positioning,arrows,trees,shapes,fit,shadows}

\usepackage{amsfonts}
\usepackage{ifthen}
\usepackage{booktabs}
\usepackage{amssymb}
\usepackage{supertabular}
\usepackage{array}
\usepackage{multirow}
\usepackage{fancyhdr}
\usepackage[normalem]{ulem}

\usepackage{amssymb}
\usepackage{supertabular}
\usepackage{array}
\usepackage{multirow}
\usepackage{fancyhdr}
\usepackage{psfrag}
\usepackage{amsmath}
\usepackage{hhline}
\usepackage{type1cm}
\usepackage{lettrine}
\usepackage[colorinlistoftodos,prependcaption,textsize=tiny]{todonotes}

\usepackage{algorithm}
\usepackage[noend]{algpseudocode}

\usepackage{xspace}

\usepackage{babel}
\usepackage[export]{adjustbox}
\usepackage{cleveref}

\usepackage{pifont}

\usepackage{scrextend}

\crefformat{section}{\S#2#1#3} 
\crefformat{subsection}{\S#2#1#3}
\crefformat{subsubsection}{\S#2#1#3}

\input{math_commands.tex}

\input{macro.tex}

\usepackage{float}

\DeclareCaptionLabelFormat{andtable}{#1~#2  \&  \tablename~\thetable}

\usepackage{tikz-cd}
\usepackage{mathtools}
\usepackage{todonotes}
\usepackage{xcolor}

\definecolor{darkblue}{rgb}{0, 0, 0.5}
\hypersetup{
    colorlinks=true,
    linkcolor=darkblue,
    citecolor=darkblue,
    pdfpagemode=FullScreen,
    }

\newcommand{\ourmodel}{{SFR-DR}\xspace}

\title{SFR-DeepResearch: Towards Effective Reinforcement Learning for Autonomously Reasoning Single Agents}

\author{%
    Xuan-Phi Nguyen\thanks{Project lead \& corresponding authors: \href{mailto:xnguyen@salesforce.com}{\{xnguyen,cxiong,sjoty\}@salesforce.com}} \qquad Shrey Pandit \qquad Revanth Gangi Reddy\thanks{Work done during an internship at Salesforce AI Research.} \qquad Austin Xu \And Silvio Savarese \qquad Caiming Xiong$^*$ \qquad Shafiq Joty$^*$  \\\\
    Salesforce AI Research 
}

\begin{document}

\maketitle

\setcounter{footnote}{0} 

\begin{abstract}

Equipping large language models (LLMs) with complex, interleaved reasoning and tool-use capabilities has become a key focus in agentic AI research, especially with recent advances in reasoning-oriented (``thinking'') models. Such capabilities are key to unlocking a number of important applications. One such application is Deep Research (DR), which requires extensive search and reasoning over many sources. Our work in this paper focuses on the development of native \textbf{Autonomous Single-Agent} models for DR featuring minimal web crawling and Python tool integration. Unlike multi-agent systems, where agents take up pre-defined roles and are told what to do at each step in a static workflow, an autonomous single-agent determines its next action dynamically based on context, without manual directive. While prior work has proposed training recipes for base or instruction-tuned LLMs, we focus on continual reinforcement learning (RL) of reasoning-optimized models to further enhance agentic skills while preserving reasoning ability. 
Towards this end, we propose a simple RL recipe with entirely synthetic data, which we apply to various open-source LLMs. Our best variant, \ourmodel-20B, achieves up to 28.7\% on Humanity's Last Exam benchmark. 
In addition, we conduct key analysis experiments to provide more insights into our methodologies.


\end{abstract}

\input{intro}

\input{background}

\input{method}

\input{experiments}

\section{Conclusion}
Overall, we propose a compact synthetic-data reinforcement learning recipe that adapts reasoning-optimized LLMs into native Autonomous Single-Agent systems for Deep Research. Applied to open-source backbones, our best variant attains 28.7\% on Humanity’s Last Exam. Our analysis highlights the contributions of various components of our system as well as impact of our RL training process on the agent behavior.

\bibliography{neurips_2023.bib}
\bibliographystyle{plainnat}


\end{document}

%% file: math_commands.tex
\usepackage{amsmath,amsfonts,bm}

\def\eqref#1{equation~\ref{#1}}

\def\1{\bm{1}}

\makeatletter
\newcommand{\xRightarrow}[2][]{\ext@arrow 0359\Rightarrowfill@{#1}{#2}}
\makeatother

\DeclareMathAlphabet{\mathsfit}{\encodingdefault}{\sfdefault}{m}{sl}
\SetMathAlphabet{\mathsfit}{bold}{\encodingdefault}{\sfdefault}{bx}{n}



%% file: macro.tex
\newcommand{\red}[1]{\textcolor{red}{#1}}



%% file: intro.tex
\section{Introduction}\label{sec:intro}

The ability to call functions (or tools) is a core and well-studied capability in building practical Large Language Models (LLMs) \citep{xlam_liu2024apigen,torl_li2025torl}. By enabling models to interact with the real world—retrieving up-to-date, reliable information through internet search or executing complex computations via code—tool use reduces hallucination and improves reliability on complex, long-horizon tasks. Among the many types of tool-integrated agents, web-based research agents, often referred to as Deep Research (DR) \citep{openai2025deepresearch}, have attracted notable attention in both closed- and open-source communities. These agents use browsing and coding tools to answer challenging questions. Unlike agents that need to follow largely irreversible, acyclic workflows of tool calls (e.g., computer-use \citep{computer_use_sager2025ai} or email agents), DR agents can invoke tools in a flexible order. However, this flexibility demands  advanced reasoning ability to plan and execute correct tool calls, e.g., searching for specific entities or writing Python code.

DR systems are typically implemented as either single-agent or multi-agent systems. A single-agent system, such as OpenAI's DeepResearch\citep{openai2025deepresearch} or Kimi-Researcher~\citep{kimi_researcher}, gives a single, tool-equipped LLM the user's question and allows it to autonomously perform actions like web search, webpage browsing, or coding in a multi-turn format. Apart from the initial prompt, this agent receives no external directives at intermediate steps. In contrast, a multi-agent DR system (e.g., OpenManus \cite{openmanus2025}, Open DR \cite{langchain_open_deep_research}) typically employs a complex workflow in which multiple agents—potentially powered by different LLMs—are assigned distinct roles and task descriptions (e.g., orchestrator, planner, coder, researcher, supervisor). For instance, an orchestrator may decompose a complex problem into sub-problems and dispatch them to specialized agents equipped with dedicated tools.

In this work, we train autonomous single-agent LLMs to perform complex tasks using a minimal set of tools: a web search tool, a web crawling tool, and a Python interpreter. These agents are trained to process initial requirements and contextual information from tool outputs to autonomously plan and execute their next action. Our focus on single-agent systems stems from two beliefs: First, we believe single agents can \textit{generalize} better to unseen tasks, as they are not constrained by the predefined, heuristic-based workflows common in multi-agent systems. Second, if more complex multi-agent scaffolding is used, single-agents can be seamlessly integrated as specialized sub-agents, thereby reducing overall system complexity by eliminating redundant deep research components. While some prior work has suggested to begin agentic training from instruction-tuned (SFT) or base (pre-SFT) models with cold-start instruction-tuning and RL \citep{tao2025webshaper,li2025websailor}, we focus on continual RL training on reasoning-optimized models \citep{qwq32b,yang2025qwen3,agarwal2025gpt} to further improve agentic capabilities while also preserving their strong reasoning ability. Towards this end, we develop a generic RL-based framework to improve the agentic capabilities of pre-trained reasoning models. The framework spans two main aspects:

\begin{itemize}[leftmargin=*]
    
    \item \textbf{Agentic Inference Pipeline}: We implement an agentic scaffolding that mirrors the way the initial LLMs are originally trained, closely resembling a multi-turn conversation with tools. Additionally, we develop a memory management system that allows the agent to manage its own memory, effectively enabling a virtually unlimited context window. Specifically, we reserve a portion of the fixed context length as a memory buffer and provide the model with a memory \emph{clean-up} tool. During roll-out, the agent will be told if the memory exceeds the token buffer length, and tasked with using the clean-up tool to select only the information it deems important.  We detail how this generic method is adapted for different base LLMs to account for their specific characteristics.

    \item \textbf{RL Training Recipe}: We developed a novel pipeline to synthesize complex search- and reasoning-intensive training datasets suitable for end-to-end RL, which are more difficult than existing open-source datasets \citep{hotpotqa_yang2018,wikimultihopqa_constructing_ho2020} and challenge even state-of-the-art DR agents \citep{openai2025deepresearch}. To make use of such data, we developed a reinforcement learning algorithm based on REINFORCE~\citep{williams1992simple} with novel modifications that stabilize the policy optimization process. In particular, we find that agentic RL training can produce very diverse rollout scenarios with varied lengths (number of tool calls/turns). To mitigate these instabilities, we propose temporal advantage normalization and strategic trajectory filtering.
\end{itemize}

We applied our recipes on three distinct reasoning models: QwQ-32B~\citep{qwq32b,qwen2.5}, Qwen3-8B~\citep{yang2025qwen3}, and the recent gpt-oss-20b \citep{agarwal2025gpt}. In the experiments, we demonstrate that our RL tuned models outperform many popular baselines of similar sizes. Notably, our best variant, trained from gpt-oss-20b, achieves up to 28.7\% in the full text-only benchmark of Humanity's Last Exam \citep{phan2025humanity}, as well as competitive scores in FRAMES \citep{frames_krishna2024fact} and GAIA\citep{gaia_mialon2023gaia}. We additionally analyze the impact of our proposed agentic workflow and the importance of length normalization in preventing degenerate tool-calling trajectories in our ablation study.

%% file: background.tex
\section{Background}\label{sec:background}

\subsection{Deep Research Agents}\label{sec:ds_agents}


Providing LLMs with Internet access to generate grounded answers for complex fact-seeking questions has been an important research topic \citep{reddy2024infogent,chen2024mindsearch}, closely related to retrieval-augmented generation \citep{ragsurvey_arslan2024survey,nguyen2024sfr}. Recent web-based agentic systems increasingly leverage reasoning-oriented LLMs \citep{dsr1_guo2025deepseek}, such as Search-R1 \citep{search_r1_jin2025search,search_o1_li2025search}. Deep Research with o3 \citep{openai2025deepresearch} was among the first systems to integrate both web browsing and code interpretation to tackle harder problems, including not only challenging short-form questions but also abstract report-writing queries. Since then, a variety of deep research systems have been introduced \citep{deepresearch_survey_xu2025comprehensive,java2025characterizing_dr,sfr_deepsearch_choubey2025benchmarking}. While some systems remain proprietary \citep{openai2025deepresearch,gemini_deep_research,perplexity_deep_research} with undisclosed architectures, many open-source systems can be broadly categorized as either single-agent or multi-agent. The boundary between the two is blurred and still debated.

In this paper, we regard a single-agent system as one where a single primary LLM autonomously makes tool-calling decisions based on the current context, without external directives at intermediate steps. Single-agent systems typically restrict themselves to primitive tools but not LLM-enhanced tools. Examples include basic LLMs with function calling \citep{yang2025qwen3,gpt5,agarwal2025gpt}, ReAct-style agents \citep{react_yao2022react,li2025websailor,tao2025webshaper}, or a repetitive tool-calling-then-summarizing process such as Kimi-Researcher \citep{kimi_researcher}.

By contrast, a multi-agent system involves multiple agents collaborating within a pre-defined workflow, each with its own agency and specialized role (e.g., planner, reasoner, coder, researcher, writer). In addition, certain tools are treated as sub-agents when they rely on LLMs to perform complex tasks, such as query-focused information extraction from web content. Many open-source deep research systems adopt this paradigm, including Open Deep Research \citep{alzubi2025opendeepsearch,langchain_open_deep_research}, Miroflow \citep{miroflow}, and others \citep{zhang2025agentorchestra,li2025webthinker,openmanus2025,alzubi2025opendeepsearch}.

Overall, single-agent systems are simple, agile, and more likely to generalize to unseen tasks {because of its flexible design and autonomous workflow}, whereas multi-agent systems tend to specialize in pre-defined workflows and can achieve greater performance and efficiency. Single-agents can also be seamlessly plugged in an multi-agent system.






\subsection{Agentic Tool-integrated Reinforcement Learning}\label{sec:tool_integrated_rl}

Reinforcement learning with verifiable rewards (RLVR) has become the standard methodology for training agentic LLMs \citep{openai2025deepresearch,gpt5,yang2025qwen3,kimi_researcher,team2025kimi}. While various RL algorithms, such as GRPO and related methods \citep{grpo_shao2024deepseekmath,dapo_yu2025dapo,gspo_zheng2025group,rloo_ahmadian2024back}, have achieved success on single-turn reasoning tasks, they have not proven to be optimal or stable for long-horizon agentic tasks without modifications. In the realm of math solving with code, techniques like trajectory filtering have been introduced to stabilize multi-turn training \citep{torl_li2025torl,xue2025simpletir,feng2025retool,arpo_dong2025agentic}. For web agents, cold-start instruction-tuning followed by RL with customized objectives and roll-out strategies is commonly employed \citep{tao2025webshaper,li2025websailor,Agent-R1,wei2025webagentr1,kimi_researcher}. Other approaches \citep{li2025webthinker,miroflow} leverage preference learning \citep{dpo_rafailov2023direct}. Most prior work begins training from base (pre-SFT) or instruction-tuned models. In contrast, we initialize from reasoning-optimized “thinking” models, allowing us to both leverage and preserve their strong step-by-step reasoning while endowing them with new agentic capabilities. This design choice, however, introduces unique challenges, most notably, the difficulty of reliably steering long-horizon chain-of-thought behavior during optimization.





%% file: method.tex
\section{SFR-DeepResearch}\label{sec:method}

In this section, we describe the process of building SFR-DeepResearch (\ourmodel). First, we formulate a novel yet simple agentic inference scaffolding that stabilizes multi-step rollouts with a flexible contextual memory buffer that enables arbitrarily length trajectories.
Then, we describe how the training data is constructed for end-to-end RL training. Finally, we explain our novel RL recipe, which helps stabilize multi-turn RL in complex deep research setups.

\subsection{Agentic Inference Scaffolding}\label{sec:agentic_scaffolding}

\subsubsection{Tools}

The tool-set available to an agentic system plays a critical role in the system performance. There are many advanced tools in existence that can make the agent's job easier, from more thorough search engines to interactive web-browsing tools and even LLM-enhanced tools (i.e., sub-agents) \citep{zhang2025agentorchestra,openmanus2025,miroflow,computer_use_sager2025ai}. However, when training a single-agent, the tools provided to the agent define a crucial component of the RL \textit{environment}. By giving an agent tools that make information extraction too easy, the agent is not challenged enough during training. For this reason, we equip and train our agents with a minimal set of tools that provide enough functionality to complete tasks, but do not render any training tasks trivially easy, incentivizing our agents to explore more and to use their tools effectively and efficiently. Concretely, we use the following three tools:

\begin{itemize}[leftmargin=*]
    \item \texttt{search\_internet(query:str)} utilizes a bare-bone search API, such as \href{https://serper.dev/}{serper.dev}, and returns the top-10 organic search results. Each result entry usually, but not always, contains the full URL, title, and a short description of the page.
    \item \texttt{browse\_page(url:str, section\_id:int)} obtains readable content from the URL by scraping and converting the raw HTML into Markdown format. Notably, we strip all hyperlinks from the HTML content, as they typically render as extremely long strings. This makes the page \textbf{unclickable} and static. As a result, the only way for the agent to discover new URLs is through the search engine. This is different from existing work \citep{openai2025deepresearch,agarwal2025gpt,li2025websailor,tao2025webshaper,li2025webthinker}, where agents can interactively browse the content of web pages and click on hyperlinks directly.
    Additionally, if a web page is longer than a pre-defined limit, it will be split into sections, which the agent may ``scroll'' to via the \texttt{section\_id} argument.
    \item \texttt{code\_interpreter(code:str)} locally executes \textbf{stateless} model-written Python code on the local machine, and times out after 5 minutes. Different from stateful interpreters \citep{openai2025deepresearch}, each stateless execution is an independent runtime and does not share variables from past executions. Notably, unlike remote cloud-based sandbox environments, our coding tool does not have access to the command line or ability to install packages. File system access and sensitive packages that pose security and integrity risks are also prohibited.\footnote{Attempts to import packages like \texttt{os,sys,subprocess,socket,signal,multiprocessing, threading,ssl,pdb,resource,xmlrpc}, etc, will receive an error message.}
\end{itemize}


\subsubsection{Single-agent Agentic Workflow} \label{sec:single}

Training function-calling LLMs is typically done by utilizing a multi-turn structure: Tool descriptions are inserted in the model system prompt, the model outputs its tool calling action in the assistant turn, and the tool results are presented to the model in either the user turn~\citep{yang2025qwen3} or specialized turns designed for tool results~\citep{agarwal2025gpt,nguyen2024sfr}. As such, when continually training models with RL to excel as DR agents, it is intuitive to preserve the multi-turn nature of function-calling training. However, we find that specific model families have distinct model characteristics that do not allow for a one-size-fit-all training formulation. As a result, we tailor our inference setups on a per-model family basis. 

\begin{figure}
\centering
\includegraphics[width=\textwidth]{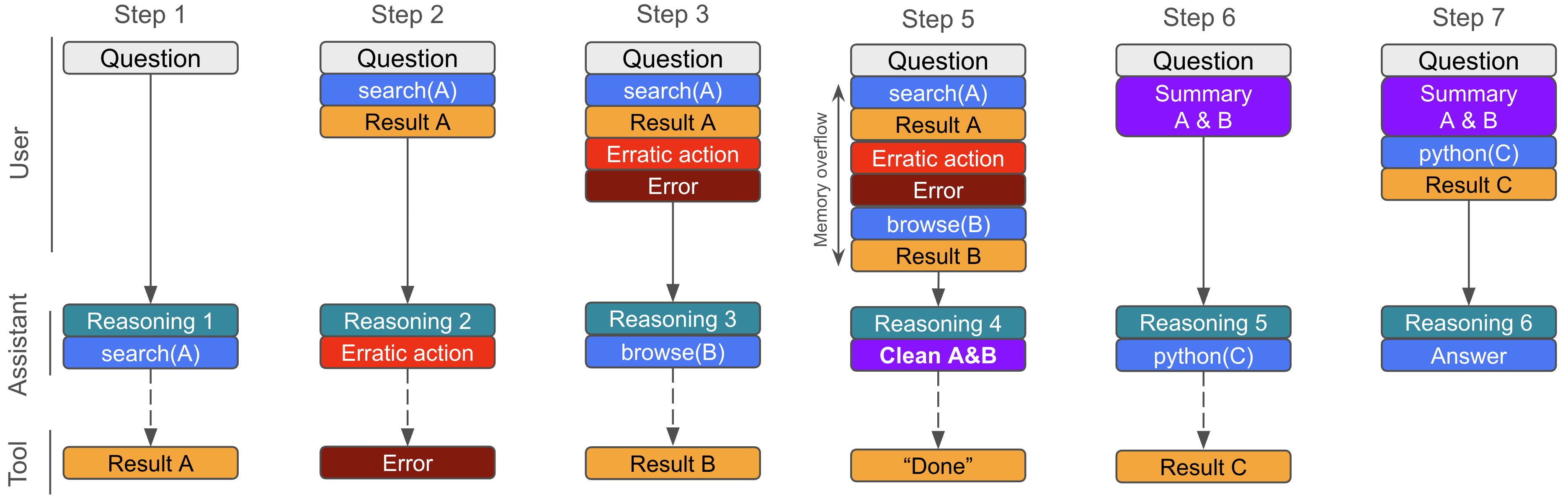}
\caption{An example tool calling trajectory by our \ourmodel agentic workflow, catered for QwQ-32B and Qwen3 models. The multi-turn interaction is framed as a single-turn contextual question answering problem, where there is always only 1 user turn. The previous tool calls and responses are packed as memory and placed in the user turn together with the question.}
\label{fig:agentic_pipeline}
\end{figure}

\Cref{fig:agentic_pipeline} demonstrates an example tool calling trajectory of our \ourmodel. Formally, given an input question $q$, a multi-turn agentic trajectory is defined as an ordered set $\mathcal{T}(q) = (q, c_1, o_1, c_2, o_2, ...,c_n, o_n, a)$ where $c_i$ is the model response that includes a tool call, $o_i$ is the resulting feedback returned from executing that tool call in the environment at agentic step $i$, and $a$ is model's proposed answer. For reasoning models, $c_i$ and $a$ may not only include the output part, but also the ``thinking'' tokens.

For \textbf{QwQ-32B \citep{qwq32b} and Qwen3 \citep{yang2025qwen3}} models, 
we find that inference is more effective when we reformulate a typical multi-turn tool calling conversation as an iterative single-turn task (Figure \ref{fig:agentic_pipeline}). We hypothesize that these models were extensively post-trained with RL to excel at typically single-step tasks, such as solving math problems or writing code. More precisely, with the default multi-turn chat template at step $i>1$, the prompt sent to the model is equivalent to ``\texttt{<user>\red{$q$}<assistant>\red{$c_1$}<user>\red{$o_1$}...<user>\red{$o_{i-1}$}<assistant>}''.
\footnote{
    For brevity, we condense the template in main text. The actual template is \texttt{<|im\_start|>user\textbackslash n\red{$q$}<|im\_end|><|im\_start|>assistant\textbackslash n\red{$c_1$}<|im\_end|>}\ \texttt{<|im\_start|>user\textbackslash n<tool\_response>\red{$o_1$}</tool\_response><|im\_end|>}...
} In our agentic workflow, the same prompt can be reformulated a single-turn contextual question answering prompt as ``\texttt{<user>\red{$q;[c_1,o_1,...,o_{o-1}]$}<assistant>}'', where the question, tool call actions, and tool results all reside in the first user turn. The tool results are included in the user turn as contextual knowledge, which the model should use to determine its next action. As more tool calls and tool results accumulate during rollout, instead of a longer multi-turn conversation, the prompt always contains only one ``contextual question'' that keeps getting longer and more complex.
Additionally, as we are training reasoning LLMs, we must manage the long CoT ``thinking'' tokens. We find that the most straightforward approach, interleaving the long CoT between tool calls, as recommended by Qwen3, causes several problems. First, the long CoTs include unnecessary tokens, which quickly fills up the model's context window. Second, we observe that the long CoT tokens begin to devolve into highly erratic outputs as the trajectory length grows, likely because existing post-training does not extend conversations to extremely long, multi-turn settings. Instead, we opt to omit previous long CoTs in the current step. As a result, at each step, the model will generate a new long CoT that reasons about the latest contextual information. We analyze the impact of this modification in \Cref{{sec:aly:agentic_pipeline}}.

On the other hand, the \textbf{gpt-oss} model \citep{agarwal2025gpt} exhibits significantly stronger multi-turn abilities and consistently produces shorter chain-of-thoughts. Therefore, we follow the default \textit{harmony} chat template that comes with the model instead of the above single-turn proposal. 

\textbf{Long-horizon Context Management.} For complex problems, the long-context nature of multi-turn tool-calling interactions becomes a critical challenge: Naively storing all tool results and model reasoning traces quickly fills up the model's context window (of length $L$ tokens), whereas blindly truncating the conversation may inadvertently discard crucial information found in earlier conversation turns. As such, context management becomes a crucial skill for DR agents, especially for models that output long CoT. Several multi-agent workflows have proposed using an external memory bank \citep{reddy2024infogent,openmanus2025}. Instead, we train our agents to self-manage its own internal memory, i.e., \textit{the context window itself}. For QwQ and Qwen models, we simply provide an internal tool called \texttt{clean\_memory(content:str)} that replaces the current context information, such as $ [c_1,o_1...,c_o,r_o]$ for step $i$, with model-generated content specified in the \texttt{content} argument.\footnote{If the agent has cleaned up K times, the current memory will not include tool results $o_j$ that occurred before the K-th clean up as they were erased.} This clean up process is illustrated in step 5 of \Cref{fig:agentic_pipeline}.
During rollout, if the memory is about to exceed a pre-defined limit $L_{\text{mem}} < L$, the model will be informed so and instructed to invoke \texttt{clean\_memory}. In this case, any other tool call other than cleaning up the memory will yield a ``memory overflow'' error until the model successfully cleans the context memory. For gpt-oss models, as we use the original multi-turn format, we instead provide them with tools to edit or delete individual past tool results.

\textbf{Improving Fault Tolerance.} As all LLMs are stochastic, there is a chance that the models will produce responses in incorrect formats, causing parsing errors. Depending on the severity of the format error, we implement protocols to either repair, retry, or inform the model about the error in the next step. For example, if the model produces tool calling action but misplaces a special token and a deterministic reparation is unsuccessful, the model will receive a syntax-error message and will be asked to remedy its tool call in the next turn, as shown step 3 of \Cref{fig:agentic_pipeline}. The model would also receive a similar warning if it calls a non-existent tool or an existing tool with invalid parameters.

\subsection{Training Data Synthesis Pipeline}\label{sec:synthetic_data_pipeline}

We create a mix of challenging synthetic data to train our agent to perform two core tasks: Short-form QA and long-form report writing. For short-form QA, we observe that existing multi-hop training datasets \citep{wikimultihopqa_constructing_ho2020,hotpotqa_yang2018} are not sufficiently challenging for our initial models, even without search; such questions are likely too easy and/or already included in pre- or post-training data mixes. 
As such, we utilize an iterative approach to construct more challenging multi-hop question-answer pairs progressively. In addition to multi-hop fact-seeking questions, we also include traditional math and code reasoning tasks.
For long-form reports, we prompt an LLM to create both instructions and grading rubrics for a curated set of open-ended questions. In all, our training data is challenging for even state-of-the-art agents: OpenAI Deep Research with o3~\citep{openai2025deepresearch} achieves less than 65\% accuracy on our short-form dataset, while our best baseline agent scored less than 40\%. These questions are also search intensive, with an o4-mini agent \citep{o4mini} taking up to 50 tool calls to complete a single question.

\subsection{End-to-End Reinforcement Learning Recipe}\label{sec:end_to_end_rl}

\textbf{Length-normalized RL Objective.} We employ a modern variant of the REINFORCE algorithm \citep{williams1992simple} to train our \ourmodel agents, which features clipped-surrogate losses \citep{ppo_schulman2017proximal,grpo_shao2024deepseekmath,gspo_zheng2025group}. Particularly, given an  input question $q$, we perform a group of $G$ independent rollout trajectories defined as the Markov chain $\tau_i = [(s_{i,1},a_{i,1}),(s_{i,2},a_{i,2}),...,(s_{i,T_i},a_{i,T_i})]$ of length $T_i$ where $s_{i,j}$ is the Markov state and $a_{i,j}$ is the corresponding policy action at agentic step $i$.
All steps $s_{i,j}$ receive the same reward $r_i$ which is obtained at the last step. 
We do not use any intermediate rewards. The step-level advantage $A_{i,j}$ is then determined as:
\begin{equation}
    A_{i,j} = A_i = \frac{r_i - \text{mean}(\overline{R})}{\text{std}(\overline{R})\cdot T_i} \label{eqn:advantage}
\end{equation}
where $\overline{R}$ is the set of all trajectory-level rewards within the same group. The loss is then computed at step level. Compared to 
prior work \citep{grpo_shao2024deepseekmath,dapo_yu2025dapo}, our advantage formulation features a normalization term over the trajectory length. This means that steps of longer rollouts have lower absolute advantage magnitudes than those of shorter rollouts. The normalization is designed to regulate and stabilize training when the trajectory length varies significantly. 
Without normalization, we observe that longer trajectories dominate the training loss, even if said trajectories are not necessarily of higher quality or correct. 
As a result, failing-but-long trajectories appear more frequently over time, even if they are imposed with penalties and negative rewards.
This results in unintended negative learned behavior, such as repeated tool calling; We find this behavior is mitigated with length normalization (\Cref{sec:aly:tool_calls}).

\textbf{Trajectory Filtering.} We also employ a strategic trajectory filtering procedure to ensure the training batch mixture is optimal. Particularly, similar to \cite{dapo_yu2025dapo,kimi_researcher}, we filter out invalid trajectories, such as those ending with truncation or incorrectly-formatted responses, from the replay buffer.
We also maintain the ratio of positive to negative trajectories of a group within a pre-defined range by randomly or heuristically dropping over-represented trajectories. We observe that, without such measures, the training process is unstable and prone to model collapse.

\textbf{Partial Rollouts.} As errors compound the longer the trajectory becomes, we found it is beneficial to reuse partial rollouts. Unlike \citet{kimi_researcher} which continues the unfinished rollouts with the updated policy, we treat the partial rollouts as new independent initial states, from which group-level Monte-Carlo rollouts begin using the same policy. This allows more exposure to and gradient updates from long-tail intermediate states.

\textbf{Reward Modeling.} For both types of tasks described in \Cref{sec:synthetic_data_pipeline}, we use the same baseline LLM as verifier but with different rewarding procedures. For short-form tasks, we prompt the LLM to decide if the agent's answer is semantically consistent with the ground-truth answer. The agent receives a reward $r_i = 1$ if the answer is consistent and $0$ otherwise. Meanwhile, for long-form report writing tasks, we employ a multi-stage process where the verifier grades the generated report individually and produces scores for multiple criteria, such as factuality/hallucination, compliance, writing quality, and citation quality. Each category carries a specific scoring weight. Within a group, valid reports are then ranked by the verifier and a ranking score can be derived for each report. The total reward of a trajectory is the weighted sum of component scores and the ranking score.

\textbf{RL Infrastructure.} We build an in-house RL training pipeline that emphasizes scalability and fault-tolerance. First, during rollout phase, the pipeline aggressively uses asynchronous and multi-processing programs to accelerate parallel rollouts, maximizing the throughput of the SGLang~\citep{zheng2024sglang} inference engine. Second, besides the search API, the toolbox is completely local and does not ping dedicated external services like \href{https://e2b.dev/}{e2b.dev}. Tool actions are all executed locally through a revamped approach that supports large-scale parallel execution. We also cache tool results so repeated tool-calls will instead retrieve results from our local database, rather than repeating the time-consuming executions. 
We further optimize the model placement of our framework, co-locating policy inference engines, verifier models, and policy models under training on the same set of GPUs to eliminate idle GPU usage. This co-location approach makes better use of GPU resources and improves training speed compared to other frameworks where inference models and training models are allocated to different GPUs. Specifically, during the gradient update phase, the inference engines are offloaded to reserve space for training models, and vice versa. 
Toolbox failures and timeouts, CPU and GPU out-of-memory (OOM) errors, hanging or crashed inference engines may disrupt ongoing RL training. To mitigate the damage of failed sub-components, we implement a recovery procedure that remedies OOM errors and restores failed components, lessening our need for human oversight during training.

\textbf{Contamination Prevention.} During both training and \textbf{evaluation} (\Cref{sec:eval}), we adopt various measures to prevent the agent from accessing potentially contaminated content or benchmark solutions via web browsing. For example, certain sensitive domain names, such as \href{hf.co}{huggingface.co}, are blocked and automatically return a ``Unavailable'' error message to the agent if access is attempted.

%% file: experiments.tex
\section{Experiments}\label{sec:eval}

%

In this section, we demonstrate the performance of \ourmodel agents across reasoning- and browsing-based benchmarks. We also discuss various ablation studies to provide more insights into the components of our method.

\subsection{Main Evaluation}\label{sec:eval_main}


\textbf{Benchmarks and Baselines.} 
We conduct experiments on three benchmarks: FRAMES \citep{frames_krishna2024fact}, GAIA \citep{gaia_mialon2023gaia}, and Humanity's Last Exam (HLE) \citep{phan2025humanity}. FRAMES and GAIA are two browsing-focused benchmarks, with the former focusing on multi-hop reasoning QA and the latter focusing on general assistant-like tasks. HLE is reasoning focused, covering domains like math and science. For GAIA, we test on the text-only evaluation set. For HLE, we conducted evaluation on the full text only subset, which includes more than 2100 questions. As some baselines have reported results on a specific subset of 500 samples \citep{li2025webthinker}, we also report on this subset as HLE-500.


\begin{table}[t]
  \caption{Performance of \ourmodel agents across benchmarks with Pass@1. Scores for our agents are obtained \textbf{under a contamination blocklist}. $\dagger$ indicates reported numbers where both the provided open-source code (if available) and publication do not mention or include leakage prevention efforts, as of September 1, 2025. $^*$ indicates baselines that we run using open-source code and our blocklist.
  }
 \label{table:eval:main_table}
  \centering
  \begin{tabular}{ll|lll}
    \toprule
    {\bf Agent} & {\bf Base models} & {\bf FRAMES} & {\bf GAIA} & {\bf HLE/HLE-500}  \\
    \midrule
    \multicolumn{5}{c}{\bf Proprietary Agents} \\
    \midrule
    Deep Research \citep{openai2025deepresearch} & o3 & - & 67.4 & 26.6/- \\
    GPT-5 \citep{gpt5} & GPT-5 & - & - & 35.2/-\\
    GPT-5-Pro \citep{gpt5} & GPT-5-Pro & - & - & 42.0/-\\
    o4-mini \citep{o4mini} & o4-mini & - & - & 17.7/- \\
    Kimi-researcher \citep{kimi_researcher} & Kimi-k1.5/k2 & 78.8$^\dagger$ & - & 26.9$^\dagger$/- \\    
    gpt-oss-20b \citep{agarwal2025gpt} & gpt-oss-20b & - & - & 17.3/- \\
    gpt-oss-120b \citep{agarwal2025gpt} & gpt-oss-120b & - & - & 19.0/- \\
    \midrule
    \multicolumn{5}{c}{\bf Multi Agents} \\
    \midrule
    OpenDeepSearch-R1 \citep{alzubi2025opendeepsearch} & Deepseek-R1-671B & 72.4* & - & 10.6*/- \\
    OpenDeepSearch-QwQ \citep{alzubi2025opendeepsearch} & QwQ-32B & 54.1* & - & 9.1*/- \\
    MiroThinker-8B  \citep{miroflow} & Qwen3-8B\&235B     & 64.4$^\dagger$ & 44.8$^\dagger$ & - \\
    MiroThinker-32B \citep{miroflow}  & Qwen3-32B\&235B   & 71.7$^\dagger$ & 54.1$^\dagger$ & -/11.8$^{\dagger}$  \\
    WebThinker-32B \citep{li2025webthinker} & QwQ-32B     & 35.5* & 48.5$^\dagger$ & -/15.8$^{\dagger}$ \\
    \midrule
    \multicolumn{5}{c}{\bf Single Agents} \\
    \midrule
    WebSailor-32B \citep{li2025websailor} & Qwen2.5-32B   & 69.78* & 44.0* (53.2$^\dagger$) & 10.75*/ \\
    WebShaper-32B \citep{tao2025webshaper} & QwQ-32B      & 69.42* & 48.5* (53.3$^\dagger$) & 12.23*/ \\
    AFM-32B     \citep{li2025afm_chain_of_agents} & Qwen2.5-32B & 55.3$^\dagger$ & - & -/18.0$^{\dagger}$ \\
    \midrule
    \ourmodel-8B  & Qwen3-8B & 63.3 & 41.7 & 13.2/14.0 \\
    \ourmodel-32B  & QwQ-32B & 72.0 & 52.4 & 16.2/17.1 \\
    \ourmodel-20B & gpt-oss-20b & 82.8 & 66.0 & 28.7  \\
    \bottomrule
  \end{tabular}
\end{table}

We compare \ourmodel against several proprietary systems and open-source multi-agent and single-agent systems. In particular, our proprietary baselines include OpenAI DeepResearch with o3~\citep{openai2025deepresearch}, GPT-5 and GPT-5-Pro~\citep{gpt5}, o4-mini~\citep{o4mini}, and Kimi-Researcher~\citep{kimi_researcher}. For open-source multi-agent baselines, we compare against OpenDeepSearch (DeepSeek-R1 and QwQ as agents) \citep{alzubi2025opendeepsearch}, WebThinker~\citep{li2025webthinker}, MiroThinker \citep{miroflow}. MiroThinker, or Miroflow, utilizes either Qwen3-8B or 32B as an orchestrator agent, but relies on Qwen3-235B sub-agents to perform reasoning tasks. Finally, for open-source single-agent systems, we evaluate against the 32B variant of the recent WebSailor~\citep{li2025websailor}, and WebShaper~\citep{tao2025webshaper} series, as well as AFM-32B~\citep{li2025afm_chain_of_agents}. 

\textbf{Evaluation with Contamination Blocklist.} 
Because all benchmarks we use are \textit{publicly available} on the Internet, any web agent may inadvertently attempt to visit websites where the benchmark solutions are hosted. In such contaminated scenarios, the agent may trivially answer the question without much reasoning effort as intended. Recent work \citep{han2025search} shows that up to 3.4\% of usable samples for HLE may be contaminated in this manner when running evaluation.
When conducting our evaluation, we mitigate this contamination risk by using a blocklist as described in \Cref{sec:end_to_end_rl}. Specifically, if the agent attempts to access a blocked domain name, such as \url{huggingface.co}, we return an ``Unavailable'' error message. 

Of our evaluated baselines, only OpenAI explicitly states that they have taken precautions with a blocklist \citep{openai2025deepresearch,gpt5,o4mini}. As such, we report OpenAI numbers directly, if numbers exist for a particular baseline. For all other baselines, we are unable to confirm if contamination precautions exist by checking both technical reports or source code. As a result, we re-run baselines using their publicly released implementations modified to use a similar blocklist, and report Pass@1. As a result, numbers that we report may differ from those reported in technical reports or online. We mark such evaluation runs appropriately when presenting our results.

\textbf{Main Evaluation Results.} \Cref{table:eval:main_table} shows the main evaluation results. Notably, our \ourmodel-20B single-agent outperforms various baselines of comparable sizes, even challenging potentially larger and proprietary agents such as OpenAI Deep Research with o3 \citep{openai2025deepresearch}. In particular, \ourmodel-20B excels not only at using its provided tools but also in complex reasoning settings like HLE, with 65\% relative improvement over gpt-oss-20b. This stems from a combination of better agentic workflow and effective RL fine-tuning {with our synthesized data}. Our 8B and 32B variants also demonstrate strong performance compared to open-source baselines \citep{li2025webthinker,tao2025webshaper,li2025websailor} trained from the same set of initial models.

\subsection{Analysis}\label{sec:analysis}
In this section, we conduct various ablation studies to provide more insights into our methods.

\subsubsection{Effectiveness of Modified Agentic Workflow}\label{sec:aly:agentic_pipeline}

We demonstrate the effectiveness of our single-agent agentic workflow for QwQ-32B and Qwen3-8B. In particular, we run inference using the default multi-turn chat templates provided by the models and our agentic workflow as described in~\Cref{sec:agentic_scaffolding}, which condenses all past turns into the first user turn. 

As shown in~\Cref{table:eval:inference}, by carefully managing multi-turn conversations, we see significant gains in performance, most notably a 10\% absolute increase on FRAMES for the 32B model. 
To explain how an inference-time trick can yield such improvements, we manually inspect trajectories produced by each model with and without our agentic scaffolding. We find a consistent theme: The ``thinking'' tokens generated in intermediate steps degrade in quality once past the first assistant turn, resulting in the model either prematurely giving up or producing degenerate or repetitive outputs. 
While these models have demonstrated strong performance in reasoning settings, such as math and code, such settings are typically conducted in single-step. Agentic research, on the other hand, requires long multi-turn interactions. We hypothesize that such models have been post-trained to focus more on single-turn reasoning tasks, and as a result, degenerate quickly as multi-turn tasks move out of training distribution. 
Our workflow re-casts the multi-turn interaction into a single-turn interaction, bringing the inference task closer to settings that the model originally excels at. This, in turn, yields training-free gains.

\begin{table}[b]
  \caption{Comparison between \ourmodel agentic workflow and the default multi-turn workflow specified in the model chat template.}
  \label{table:eval:inference}
  \centering
  \begin{tabular}{lcc}
    \toprule
    {\bf Agent} & {\bf FRAMES} & {\bf HLE}  \\
    \midrule
    Qwen3-8B multi-turn   & 52.5 & 8.8 \\
    QwQ-32B multi-turn    & 58.0 & 12.3\\
    \midrule
    \ourmodel-8B  (pre-RL) & 58.8 & 9.9 \\
    \ourmodel-32B (pre-RL) & 68.0 & 13.9 \\

    \bottomrule
  \end{tabular}
\end{table}

Our agentic workflow also features fault tolerance, which attempts to correct model errors that arise out of inference stochasticity. As described in \Cref{sec:agentic_scaffolding}, we attempt to repair faulty tool calls, returning error messages or retry if the repair is unsuccessful. These fault tolerance mechanisms help steer the model back on the right track if a misstep occurs at an intermediate step, allowing the model continue to run until it proposes a final answer.

\subsubsection{More Tool Calls Do Not Equate Better Scores}\label{sec:aly:tool_calls}

\begin{figure}
\centering
\includegraphics[width=0.6\textwidth]{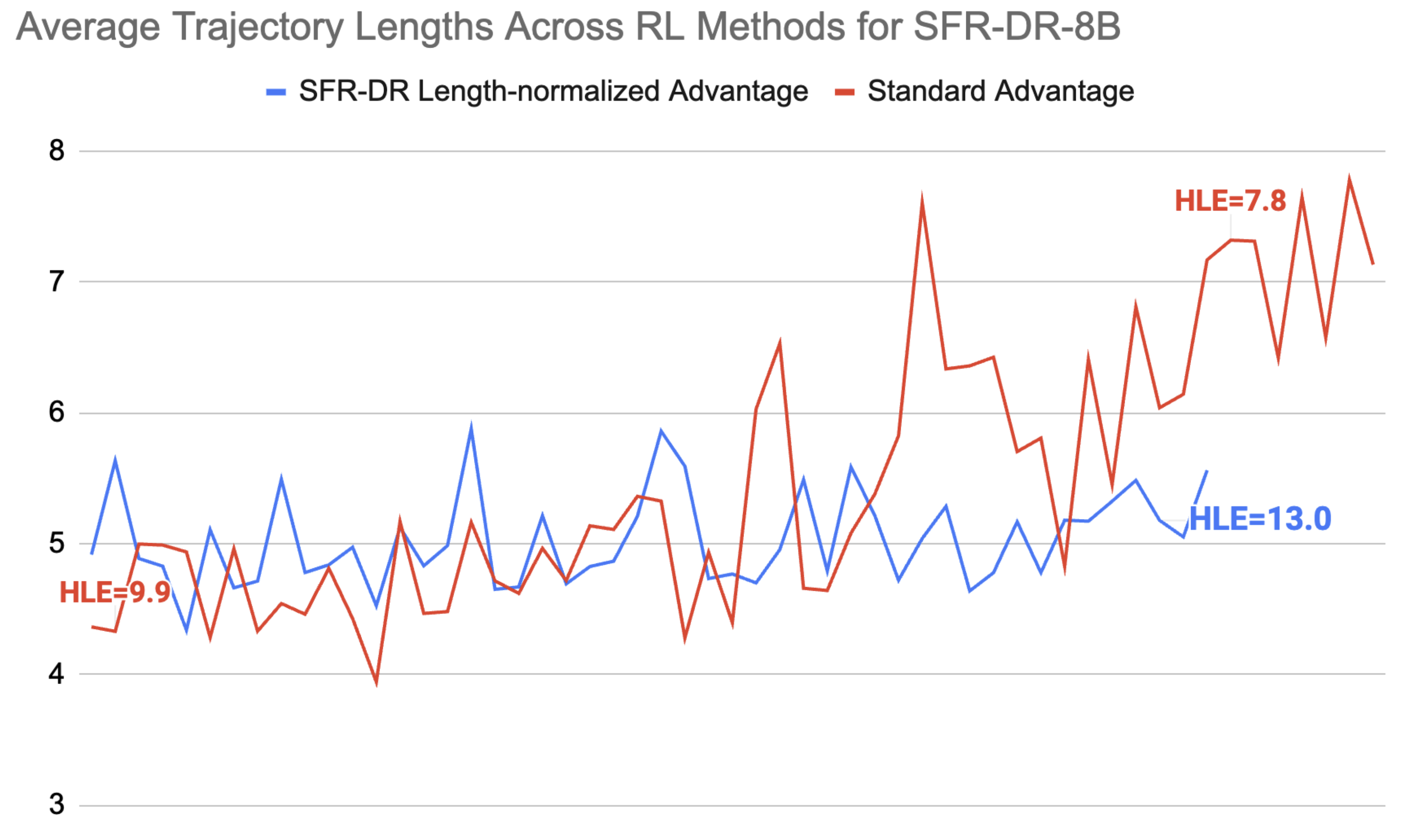}
\caption{Average training trajectory lengths of SFR-DR-8B agents over the course of RL training with and without our proposed length normalization (\Cref{sec:end_to_end_rl}). Without the normalization term, tool usage rapidly increases but the performance on HLE \citep{phan2025humanity} collapses. On the other hand, our length normalization regulates tool usage, improving performance.}
\label{fig:traj_length}
\end{figure}

Prior work suggests that increased tool usage can improve performance, framing this as a form of test-time scaling \citep{openai2025deepresearch,kimi_researcher}. This intuition holds only when tool calls are diverse and strategically executed. Under a standard RL with verifiable rewards (RLVR) objective, the assumption is that achieving the correct final answer reflects effective tool use, although this outcome is more hoped for than guaranteed. Crucially, the training objective itself does not explicitly enforce such behavior. \Cref{fig:traj_length} compares training runs with and without our length-normalized advantage modification introduced in \Cref{sec:end_to_end_rl} for SFR-DR-8B agents. Without normalization, tool usage rises rapidly, but both training reward and validation performance decline. Closer inspection shows that the agent degenerates into making repetitive, identical tool calls. This occurs because long trajectories contribute disproportionately more action steps to each batch, dominating the loss; as a result, they are reinforced and appear more frequently, even when heavily penalized through negative rewards or length penalties. In contrast, with normalization, training is more stable and performance improves, even though tool usage grows more moderately.

\subsubsection{Tool Usage Analysis}\label{sec:aly:tool_usage}

\begin{figure}[t]
  \centering
  \begin{subfigure}[t]{0.49\textwidth}
    \centering
    \includegraphics[width=\linewidth]{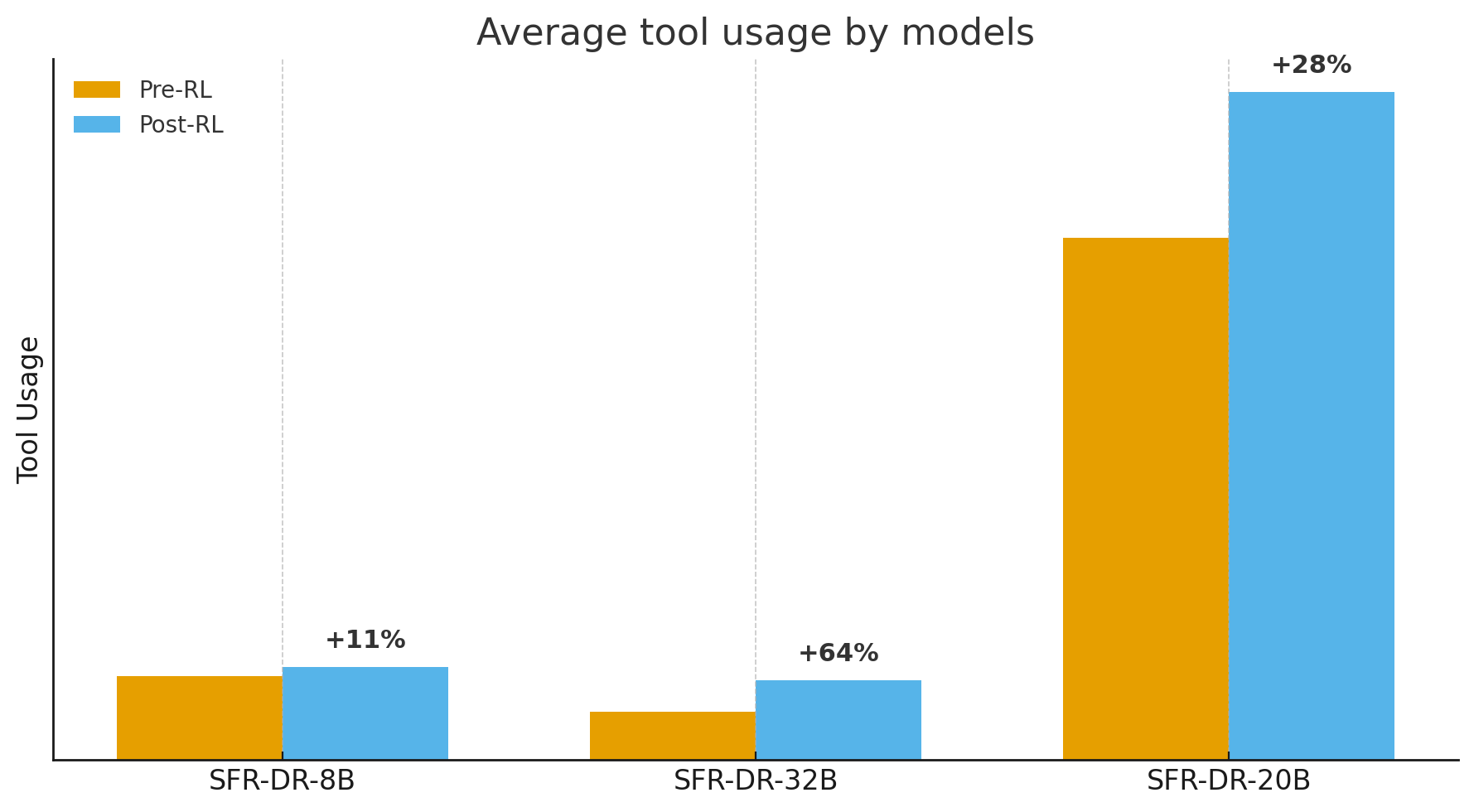}
    \subcaption{Average tool usage by models}
    \label{fig:num_tool_calls}
  \end{subfigure}\hfill
  \begin{subfigure}[t]{0.49\textwidth}
    \centering
    \includegraphics[width=\linewidth]{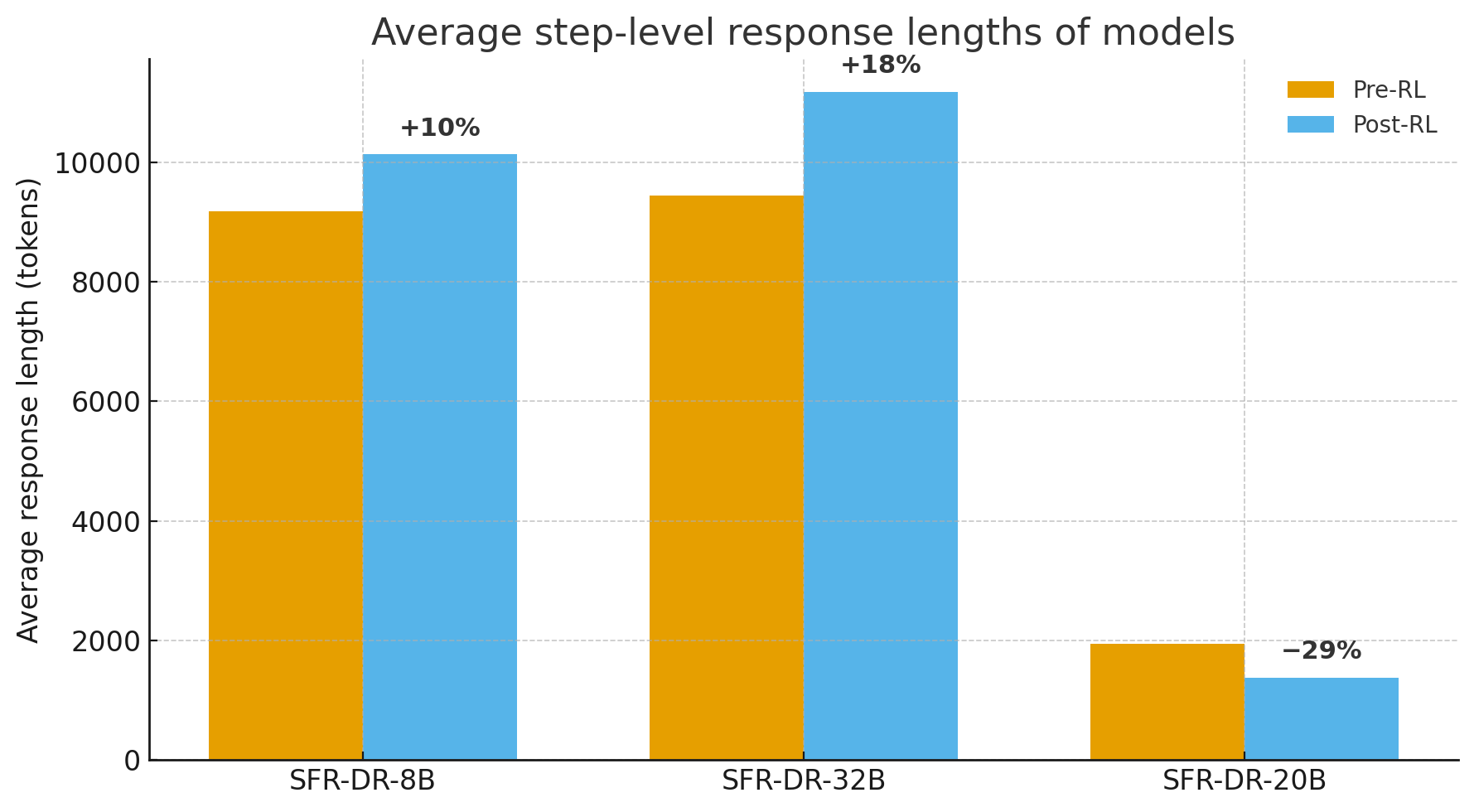}
    \subcaption{Average step-level model response length}
    \label{fig:step_level_response_lengths}
  \end{subfigure}
  \caption{Comparison of (a) average tool usage and (b) average step-level response lengths (tokens) on HLE across different models.}
  \label{fig:tool_usage_vs_lengths}
\end{figure}

\Cref{fig:num_tool_calls} reports the average number of tool calls made by different \ourmodel agents before and after RL training on HLE. As shown in~\Cref{sec:aly:tool_calls}, even though our RL objective limits rapid increases in tool usage, RL training produces models that make moderately more tool calls than pre-RL variants, improving overall performance. 
Another interesting observation is that \ourmodel-20B, which is trained from gpt-oss-20b, makes up to 10 times more tool calls than QwQ and Qwen3 variants, which tend to do only internal reasoning for many samples. This behavior may be a consequence of our hypothesis in~\Cref{sec:aly:agentic_pipeline} that the Qwen-family of models have been optimized primarily for single-turn reasoning tasks. On the other hand, gpt-oss-20b appears better primed for agentic training, as indicated by its pre-RL number of tool calls. RL training further improves this number.

\subsubsection{Response Length Analysis}\label{sec:aly:response_length}

The response length, primarily the length of ``thinking'' chain-of-thoughts (CoTs), provides another insight into model behavior. \Cref{fig:step_level_response_lengths} reports the average number of generated tokens at each agentic step (step-level) for different \ourmodel models on HLE. As shown, the \ourmodel-20B model generates less than 2,000 tokens per step, which is 4-5 times fewer than the 8B and 32B counterparts. This indicates that the gpt-oss-20b model is much more token-efficient than the Qwen-family models, with the latter tending to overthink and generate excessively long per-step CoTs. These inherently longer CoTs also make these models more difficult to steer via fine-tuning. 
More interestingly, while the RL training leads to an increase in average response length for QwQ and Qwen3 models, it causes the gpt-oss-20b variant to shrink in response length, making it even more token-efficient.